\documentclass[12pt]{article}
\usepackage[margin=1in]{geometry}
\usepackage{setspace}
\usepackage{graphicx}
\usepackage{booktabs}
\usepackage{amsmath,amssymb}
\usepackage{natbib}
\usepackage{hyperref}

\usepackage{algorithm}
\usepackage{algpseudocode}
\usepackage{tikz}
\usetikzlibrary{positioning,arrows.meta,decorations.pathreplacing,shapes.geometric,calc}
\usepackage{eurosym}
\usepackage{subcaption}
\usepackage{graphicx}
\usepackage{tabularx}
\usepackage{booktabs}
\usepackage{pdflscape}
\usepackage{enumitem}

\usepackage{natbib}
 \bibpunct[, ]{(}{)}{,}{a}{}{,}%
\begin{document}

\title{Multichoice Modeling: Exploring Decision Configuration Space under Uncertainty}
\author{%
Mariia Kozlova\textsuperscript{1,*}, Roman Stepanov\textsuperscript{1},\\
Alexander Walzer\textsuperscript{2}, Andrea Saltelli\textsuperscript{3}\\[0.7em]
\normalsize \textsuperscript{1}LUT University, LUT Business School, Lappeenranta, Finland\\
\normalsize \textsuperscript{2}University of Applied Sciences and Arts Northwestern Switzerland FHNW, Switzerland\\
\normalsize \textsuperscript{3}UPF Barcelona School of Management, Barcelona, Spain\\[0.4em]
\normalsize \textsuperscript{*}Corresponding author: mariia.kozlova@lut.fi
}
\date{}
\maketitle
\begin{abstract}

Decision support has become sophisticated at exploring uncertainty, but much less so at representing the architecture of the decision itself. As a result, the decision space is often narrowed before uncertainty is analyzed, leaving interactions among choices and the effects of alternatives they imply largely invisible. This paper introduces multichoice modeling, a framework for making large decision configuration spaces explicit and computationally explorable under uncertainty. Drawing on the logic of rugged landscapes, we conceptualize such problems as contingent performance landscapes, where interdependent choices generate configuration-specific outcomes whose value changes across uncertain futures. The computational framework combines multidimensional decision and uncertainty spaces in a single simulation. Further, variance-based sensitivity analysis is used for identifying influential choices, and conditional distribution analysis for comparing configuration-specific performance profiles. We demonstrate the framework through three case studies: a manufacturing process upgrade, M\&A strategy, and mortgage design. By contrasting conventional uncertainty-only analysis with multichoice modeling, the cases show how opening the decision configuration space can reveal unexpected complementarities, near-best configurations with different risk profiles, and strategies that were not apparent as alternatives at the outset.
\end{abstract}

\noindent\textbf{Keywords:} Decision support; uncertainty; strategy table; Monte Carlo simulation; global sensitivity analysis


\begin{quote}
\raggedleft
``\textit{We shape our tools and thereafter our tools shape us}''\\
Various authors, attributed to Marshall McLuhan \citep{Alex_Kusky}
\end{quote}

\section{Introduction}\label{sec:Intro}

In investment appraisal, cost-benefit analysis, operations, finance, and risk management, actions must be evaluated under uncertain future conditions. Therefore, the treatment of uncertainty has become a central aspect of decision support \citep{saltelli2020five,scoones2024navigating}. Yet a form of residual determinism remains in the specification of the decision architecture itself. Exploratory modeling often examines how a single system or policy behaves under uncertainty \citep{bankes1993exploratory,kwakkel2013dynamic}. Robust decision-making evaluates a limited set of candidate strategies across many plausible futures \citep{lempert2006general,marchau2019dmdu}. Real options analysis frequently focuses on valuing or optimizing either a single or a limited set of generic flexibilities \citep{dixit1994investment,trigeorgis1996real,kozlova2017real}. Consequently, decision support often remains confined to narrowly defined decision situations, in which only a small subset of the available strategic degrees of freedom is examined.


 There is an emerging call for opening up the decision space itself \citep{hamalainen2024generating,stepanov2024unlocking,colorni2020designing}. Unlike the decision analysis field, strategy research has long recognized that a strategy is not a single decision, but a configuration of multiple interdependent choices \citep{levinthal1997adaptation,rivkin2000imitation}. Drawing on the concept of rugged landscapes, this literature shows that when choices interact, the performance contribution of any choice depends on the broader configuration in which it is embedded. Yet the rugged-landscape perspective is primarily used as a behavioral and evolutionary lens in explaining how firms search and adapt, whereas strategy evaluation raises a qualitatively different decision-support problem.

Consider an example of a battery supply strategy for an electric vehicle (EV) manufacturer. The firm faces multiple interdependent choices, including whether to make or buy batteries, whether to own upstream mining assets, which chemistries to support, which suppliers and regions to rely on, which transportation routes to choose, how much inventory buffer to arrange, and whether to use long-term contracts or spot purchasing. 
These choices are interdependent and interact with each other: battery chemistry determines suppliers and production assets; vertical integration alters the value of diversification and contracting decisions; transportation, inventory, and sourcing choices jointly determine supply-chain resilience and cost. External uncertainties, such as EV demand, raw-material prices, regulation, geopolitical disruptions, and transport bottlenecks, further complicate the decision landscape by altering which configurations prove robust, fragile, or unexpectedly valuable. There could be thousands of possible combinations of choices that form distinct strategies, most of which are never evaluated or even formulated. The relevant question is therefore no longer how a given strategy performs under uncertainty, but which combinations of choices give rise to attractive strategies across multiple uncertain futures.

We conceptualize strategic decision-making under uncertainty as navigation of an \textit{uncertainty-contingent rugged landscape}: a recombinatory decision space in which the value of each choice depends on both the surrounding configuration of choices and uncertain external conditions. Choice interdependencies are not abstract assumptions, but arise directly from the structure of the problem itself: physical constraints, operational dependencies, and economic mechanisms.

We further propose \textit{multichoice modeling} as a methodological framework for making such decision-uncertainty spaces explicit and computationally explorable. The approach represents a decision problem as a set of admissible choice configurations. Choices may be independent or conditional, mutually exclusive or multiple-selection, and may include option-like alternatives. The choice configurations are simulated jointly with uncertainty realizations, producing a dataset that links choices, uncertainties, and performance outcomes. Efficient variance-based sensitivity analysis and conditional projections are then used to identify influential choices and compare the performance profiles induced by their combinations. Rather than framing the problem as the fixed set of alternatives that existing methods can handle \citep{quade1980pitfalls}, strategies can emerge as combinations of interdependent choices under uncertainty.



The remainder of the paper is organized as follows. Section 2 positions multichoice modeling within the related literature on decision-making under uncertainty. Section 3 introduces the mathematical formulation and computational implementation of multichoice modeling. Section 4 illustrates the framework through three case studies: a manufacturing process upgrade conducted i a project with Hyperion Robotics, M\&A strategy consulted by M\&A experts, and a household finance decision. Section 5 proceeds with discussion, and Section 6 concludes.

\section{Related literature}

Our work contributes to the literature on decision-making under uncertainty by combining explicit construction of the decision space with exploratory analysis. We therefore position it by examining how existing approaches represent the decision and uncertainty space and whether they optimize or explore it. 

The fundamentals of decision analysis distinguish decision alternatives (controllable choices) from uncertainty (uncontrollable circumstances) \citep{howard1988decision}. This conceptual separation has often been operationalized through a narrow decision space: in cost-benefit analysis, investment analysis, risk assessment, and multi-criteria decision-making approaches, analysts first define a limited set of alternatives and then examine how uncertainty affects their consequences \citep{boardman2018cost,morgan1990uncertainty,clemen1999making,keeney1993decisions}. 

When the decision space itself is made explicit, it is commonly represented as a feasible set over which an objective is optimized. This is the case in stochastic programming, where decisions are optimized over scenarios or probability distributions \citep{kleywegt2002sample,shapiro2009lectures}, and in simulation optimization, where stochastic simulation is embedded within a search for high-performing or optimal decisions \citep{fu2015handbook}. 
Optimization makes large decision spaces tractable, yet it treats the surrounding configuration space mainly as a feasible region to be searched rather than as an object of interpretation.

Strategy literature has conceptualized configurational complexity through the notion of rugged landscapes \citep{levinthal1997adaptation,rivkin2000imitation}. In NK-type models, the complexity of the landscape is characterized by the number of decision elements $N$ and the extent of their interdependence $K$, such that the performance contribution of each choice depends on the configuration of other choices. As interdependence increases, the landscape becomes increasingly rugged. This is precisely what makes adaptive search and optimization more difficult \citep{ganco2009nk}. The appropriate functional form for representing interdependence remains debated  \citep{rahmandad2019interdependence}. Subsequent work has extended rugged landscapes to evolving ecosystems and dynamic interdependencies \citep{ganco2020rugged}. However, these approaches rarely show how external uncertainty conditions the performance landscape itself. The primary focus remains on how actors search rugged landscapes, rather than on decision-makers' ability to interpret a recombinatory decision space in which performance changes across uncertain futures.

Several related streams of literature shift the emphasis from optimization to exploration. Exploratory modeling and analysis (EMA) uses simulation to examine system behavior across many plausible assumptions and uncertainties \citep{bankes1993exploratory,kwakkel2013dynamic}. Robust decision-making (RDM) builds on this logic by evaluating candidate strategies across multiple possible futures and identifying strategies that perform satisfactorily \citep{lempert2006general,marchau2019dmdu}. Scenario discovery provides a key computational technique in this tradition: it works backward from outcomes of interest to interpretable regions of the simulation input space associated with those outcomes \citep{lempert2008comparing,bryant2010thinking,terhorst2026rule}. Global sensitivity analysis (GSA) is similarly based on exploring simulation experiments, but its main purpose is to quantify the contribution of input variables and model structures to output variation \citep{saltelli2008global,tarantola2024annotated}. Some GSA applications extend this logic beyond parametric uncertainty by treating structural modeling choices, competing scenarios, or alternative policy-relevant representations as sampled inputs \citep{Saltelli_Tarantola_Campolongo_2000}. Despite their exploratory nature, these approaches usually treat the decision side as a single architecture, a predefined set of candidate policies, or a limited number of alternatives. Therefore, they capture the broader space of possible decision configurations only partially.

Real options theory offers a powerful way of thinking about decision design under uncertainty: flexibility embedded in the initial design can create value by enabling delayed responses to unfolding uncertainty \citep{dixit1994investment,trigeorgis1996real}. Its relevance to strategic management has been articulated through the role of flexibility, staged commitment, and strategic investment timing under uncertainty \citep{trigeorgis2017real}. In practice, however, real options models often focus on valuing or optimizing a single flexibility mechanism specified prior to the numerical analysis, such as investment timing or capacity choice \citep{alam2024sensitivity}, while the broader space of interdependent strategic choices remains at best partially explored due to methodological constraints. 


Portfolio decision analysis (PDA) moves closer to a formal treatment of large decision spaces under uncertainty by representing alternatives as portfolios of actions subject to resource and feasibility constraints \citep{salo2011invitation,liesio2021portfolio}. Recent work also allows selected actions to affect the probabilities of uncertainty scenarios \citep{vilkkumaa2025framework}. The purpose, however, remains portfolio recommendation rather than exploration of the configuration space. Moreover, the additive scenario-wise utility representation supports tractable dominance analysis but is less suited to rugged landscapes.

Multichoice modeling contributes to decision-making under uncertainty by combining three dimensions that are rarely brought together: (i) a large decision space represented through multiple actionable choices; (ii) a rich uncertainty space that both influences and is influenced by choices, jointly shaping the performance of choice configurations; and (iii) an exploratory orientation aimed at discovering performance profiles across choice configurations.

\section{Multichoice modeling}

A \emph{multichoice model} represents a decision problem under uncertainty by forming a configuration space from interdependent choices, overlaying it with uncertainty realizations, and mapping this decision-uncertainty input space onto performance metrics. The framework is exploratory: rather than identifying an optimal choice configuration, it characterizes the behavior of performance metrics across the space of feasible decision designs. The following subsections formalize the framework and describe its implementation.

\subsection{Mathematical formulation}

A multichoice model consists of a set of choices \(\mathcal{C}=\{1,\ldots,C\}\), each with alternatives \(\mathcal{A}_c\), a dependency structure \(\mathcal{G}\), an uncertainty space \(\mathcal{X}\), a performance-measure space \(\mathcal{Y}\), and a model function \(f\). Formally,
\begin{equation}
\mathcal{M} = (\mathcal{C}, \mathcal{A}, \mathcal{G}, \mathcal{X}, \mathcal{Y}, f),
\end{equation}
where \(\mathcal{G}\) governs admissible combinations of alternatives and \(f\) maps selected alternatives and uncertainty realizations onto performance outcomes.

A configuration of choices is represented by a binary vector \(z=(z_a)_{a\in\mathcal{A}}\), with \(z_a\in\{0,1\}\), where \(z_a=1\) indicates that alternative \(a\) is selected. The set of all such vectors forms the unconstrained configuration space \(\bar{\mathcal{S}}=\{0,1\}^{|\mathcal{A}|}\).

Admissibility is governed by a dependency structure \(\mathcal{G}\), defined as a directed hypergraph \(\mathcal{G} = (\mathcal{A}, \mathcal{E}),\) where each hyperedge \(e=(U,V,\phi_e)\) links a set of antecedent alternatives \(U \subseteq \mathcal{A}\) to a set of affected alternatives \(V \subseteq \mathcal{A}\) through a set-valued logical mapping \(\phi_e:\{0,1\}^{|U|}\rightarrow 2^{\{0,1\}^{|V|}}\). The mapping specifies which selections in \(V\) are admissible for each selection pattern in \(U\). The dependency structure \(\mathcal{G}\) includes both constraints across choices and selection logic within each \(\mathcal{A}_c\), such as exactly-one or multiple-selection rules. A configuration \(z\in\bar{\mathcal{S}}\) is admissible if, for every hyperedge \(e=(U,V,\phi_e)\), \(z_V\in\phi_e(z_U)\). The admissible configuration space is therefore
\begin{equation}
\begin{aligned}
\mathcal{S}
= \Big\{ z \in \bar{\mathcal{S}} :\;&
z_V \in \phi_e(z_U)
\text{ for all } e \in \mathcal{E}
\Big\}.
\end{aligned}
\end{equation}

Uncertainty is represented by a random element \(X \in \mathcal{X}\). The uncertainty space may include scalar variables \(X^{(s)}\) and time-indexed stochastic processes \(X^{(t)}=\{X_t\}_{t=1}^T\). When uncertainty exposure depends on the choice configuration, a family of conditional distributions is built
\begin{equation}
X \sim P(\cdot \mid z).
\end{equation}

Given a configuration \(z \in \mathcal{S}\) and a realization \(x \in \mathcal{X}\), performance outcomes \(Y \in \mathcal{Y} \subseteq \mathbb{R}^M\) are generated by
\begin{equation}
Y = f(z,x).
\end{equation}
Dynamic and conditional decisions can be handled either as ex ante choices included in \(z\) or as contingent operational rules embedded in \(f\). 

The simulation experiment and exploratory analysis are presented next as part of the computational framework.

\subsection{Implementation workflow}

\begin{figure}[ht]
    \centering
    \includegraphics[width=\textwidth]{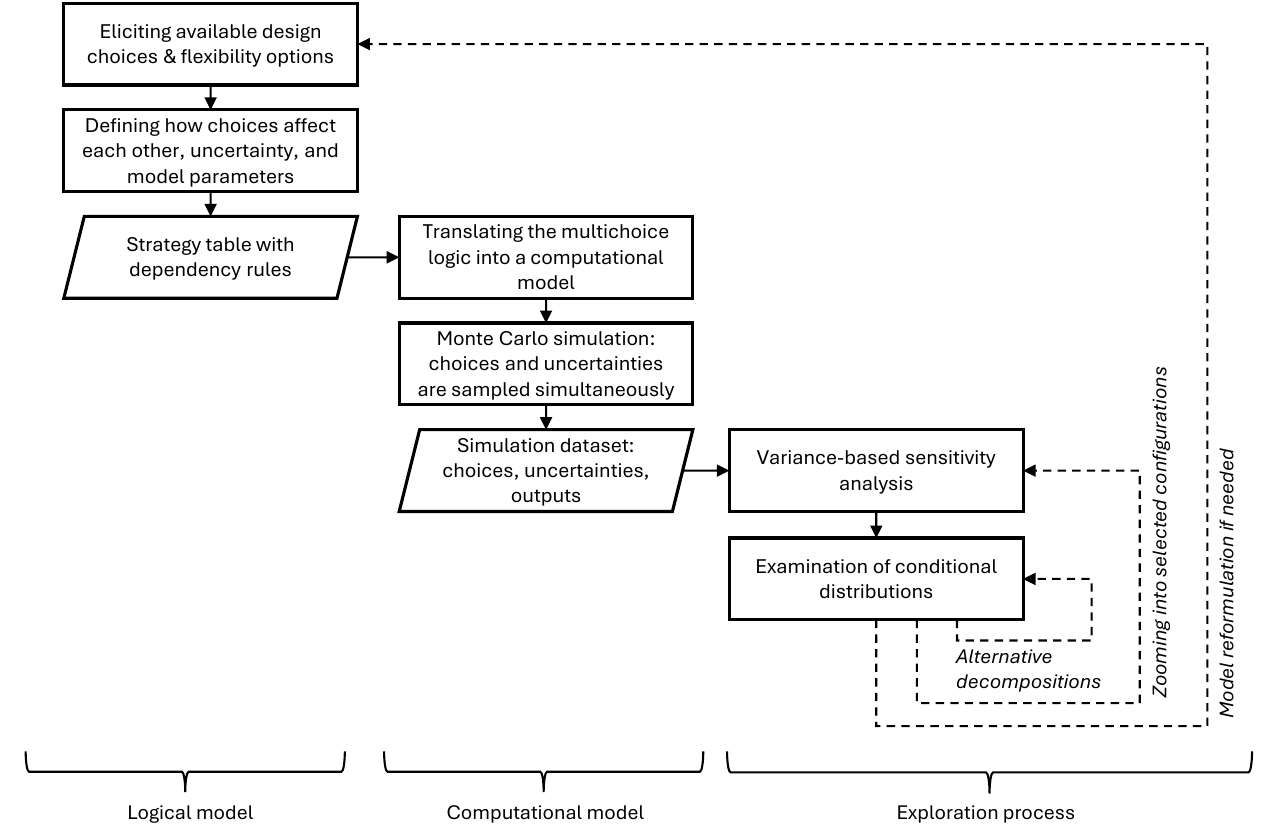}
    \caption{Implementation workflow for multichoice modeling.}
    \label{fig:mc_workflow}
\end{figure}

The implementation workflow for multichoice modeling consists of three main steps: constructing the logical model of available choices, translating it into a computational model, and exploring the resulting simulation outcomes (Figure~\ref{fig:mc_workflow}). The following subsections describe each step.

\subsubsection{Logical model: choice elicitation.}

The strategy table \citep{howard1988decision,gregory2012structured} provides the elicitation layer for multichoice modeling. Its role is to specify the decision space and the links among choices, model parameters, and uncertainty. For multichoice modeling, the table records:

\begin{itemize}
    \item \emph{Decision areas}; thematic groups of choices relevant to the decision situation. 
    
    \item \emph{Choices and alternatives}; the actions available within each decision area. 
    
    \item \emph{Choice logic}; specifying whether alternatives are mutually exclusive, allow multiple selections, introduce flexibility, or involve allocation and sizing decisions. 

    \item \emph{Dependencies among choices}; including enabling conditions, incompatibilities, sequencing requirements, and other logical or operational constraints. 
    
    \item \emph{Effects on model parameters}; such as capital expenditure, operating cost, capacity, or efficiency. When several choices affect the same parameter, their joint effect must be specified. 
    
    \item \emph{Effects on uncertainty exposure}; indicating whether a choice introduces, removes, or modifies uncertain factors. 
    
\end{itemize}

\subsubsection{Computational model and simulation.}

The strategy table is translated into the formal representation \(\mathcal{M}\) by specifying \(\mathcal{C}\), \(\mathcal{A}_c\), \(\mathcal{G}\), \(\mathcal{X}\), \(\mathcal{Y}\), and \(f\), so that the computational model reflects the elicited choice logic, parameter effects, uncertainty exposure, and dependencies.

\paragraph{Sampling and distribution types.} A direct way to examine the configuration space is to enumerate all admissible combinations of choices. This becomes impractical as the number of choices grows. Instead, the design distribution \(q\) is induced by randomly sampling alternatives for each choice uniformly, subject to the admissibility constraints encoded in \(\mathcal{G}\). This sampling scheme is not intended to represent managerial beliefs; it rather serves as a design-of-experiments mechanism that provides broad coverage of the admissible configuration space without enumerating the full factorial set of possible configurations. Uncertainty sources are sampled from the distributions specified in the model, with uniform distributions used as neutral exploratory specifications when no probabilistic information is available. Both decision configurations and uncertainty realizations vary in the generated experiment, while retaining different interpretations: \(q\) is a design distribution over configurations, whereas \(P(\cdot \mid z)\) represents uncertainty conditional on the selected configuration.

\paragraph{Sampling strategy.} Variance-based sensitivity analysis is often implemented with a quasi-random sampling strategy to improve the precision of estimated indices \citep{saltelli2008global}. However, the admissible domain in multichoice modeling is generally non-rectangular because choices may enable, exclude, or condition other choices. For such constrained decision--uncertainty domains, simple random sampling is recommended over others \citep{kucherenko2011identification}. Moreover, the simple binning estimator \citep{kozlova2025simple} used in the subsequent analysis has been shown to perform well under both simple random and quasi-random sampling, including such small sample size as $10^2$ \citep{helo2026regional}. In multichoice modeling, we therefore use simple random sampling.

\paragraph{Sample size.} In computer experiments, the required sample size is typically formulated as a function of input dimensions $d$. A widely cited rule of thumb for initial designs prescribes $N=10d$ \citep{loeppky2009choosing}. Later work emphasizes that adequate sample size also depends on response-surface complexity, effect sparsity, and the purpose of the experiment \citep{harari2018computer}. In multichoice modeling, however, sampling pressure is not determined by input dimensionality alone, because the analysis compares outcome distributions across admissible choice configurations. We therefore report sample size relative to the cardinality of the admissible configuration space \(|\mathcal{S}|\) and the number of uncertainty sources \(d_X\). As a practical benchmark, for analyses intended to compare full admissible configurations, we recommend

\begin{equation}
\label{eq:sample-size}
    N=100|\mathcal{S}|(d_X+1).
\end{equation}

Larger samples are advised when the model includes stochastic process uncertainty, tail behavior, or rarely occurring configurations induced by choice dependencies. Smaller samples may still support low-dimensional conditional analysis, but should not be interpreted as stable evidence for every full configuration.

\paragraph{Simulation dataset.} Each simulation run draws one feasible choice configuration \(z_r \in \mathcal{S}\) from a design distribution \(q(z)\) and one uncertainty realization \(x_r \sim P(\cdot \mid z_r)\), producing model outputs \(y_r=f(z_r,x_r)\):
\begin{equation}
\label{eq:dataset}
D_N = \{(z_r,x_r,y_r)\}_{r=1}^N.
\end{equation}
The resulting dataset, with choice and uncertainty variables treated as analyzed inputs \(\mathcal{W}\), forms the basis for subsequent sensitivity-guided exploratory analysis.

\subsubsection{Exploratory analysis.}

Multichoice modeling examines how configurations of interdependent choices interact with uncertainty to shape performance outcomes. Rather than identifying a single optimal configuration, the framework explores the simulated mapping \((z,X)\mapsto Y\) to identify influential choices, uncertainty conditions, and partial configurations associated with distinct outcome distributions.

\paragraph{Variance-based sensitivity analysis.} First, global sensitivity analysis is used to screen the analyzed inputs \(\mathcal{W}\), which consists of choice variables and uncertainty variables, according to their contribution to output variation. The efficient binning estimator of Sobol' indices is employed because it can be applied to dependent data and does not require a specific experimental design, unlike mainstream estimators \citep{kozlova2025simple}. For an input variable \(W_i \in \mathcal{W}\), the first-order effect measures the variance explained by conditioning on \(W_i\) alone:
\begin{equation}
S_i=\frac{\operatorname{Var}_{W_i}\left(\mathbb{E}[Y\mid W_i]\right)}{\operatorname{Var}(Y)}.
\end{equation}

For a pair \((W_i,W_j)\), the second-order effect measures the additional variance explained by their joint conditioning beyond the individual effects:
\[
S_{ij}
=
\frac{
\operatorname{Var}_{W_i,W_j}
\left(
\mathbb{E}[Y\mid W_i,W_j]
\right)}
{\operatorname{Var}(Y)}
-
S_i-S_j.
\]

Combined sensitivity indices $C_i$ aggregate these effects into an importance measure for ranking variables

\begin{equation}
    C_i = S_i + \frac{1}{2}\sum_{j \neq i} S_{ij}.
\end{equation}

In practice, often only a small subset of inputs explains the majority of output variance in computational models \citep{loeppky2009choosing}. The sensitivity results therefore serve as a navigation device: they indicate which choice combinations should be examined first.

\paragraph{Conditional outcome distributions.}
Second, the framework examines conditional outcome distributions associated with influential choices identified by the sensitivity ranking. Direct comparison of all configuration-level distributions \(P(Y\mid z)\) is typically infeasible when \(\mathcal{S}\) is combinatorially large. Instead, the analysis compares conditional projections of the form
\begin{equation}
P(Y \mid z_B = \alpha_B),
\end{equation}
where \(B\subseteq\mathcal{C}\) is a selected subset of influential choices, \(z_B\) denotes the components of \(z\) associated with the alternatives pertaining to those choices, and \(\alpha_B\) ranges over admissible partial configurations of \(B\). Thus, each conditional distribution corresponds to one admissible combination of alternatives for the selected choices, while uncertainty realizations and all non-selected choices continue to vary over their admissible ranges within the corresponding subset of the simulation dataset. The comparison therefore does not evaluate isolated deterministic strategies, but conditional projections of the full choice--uncertainty space. The algorithms for sensitivity screening and conditional distribution analysis are available as open-source implementations in Python, R, and Matlab, and through a free no-code web dashboard \citep{kozlova2024uncovering,roy2024simulation}.

\paragraph{Results interpretation.}
Choice configuration performance is assessed through the location and shape of the conditional outcome distributions. Differences in location indicate partial configurations associated with either higher or lower average performance, while differences in spread show how much variation remains after conditioning on the selected choices. The distributions also reveal choice interactions, since the effect of one choice may differ depending on the alternatives selected for other choices, rendering some alternatives attractive only in specific configurations. They can further reveal uncertainty-dependent performance, where a choice performs well only in particular regions of the uncertainty space, as opposed to configurations that remain satisfactory across a wider range of conditions. The same decomposition logic can be applied to secondary outputs to assess tradeoffs across performance dimensions.

\paragraph{Iterative analysis.} Once appealing, robust, or problematic configuration regions have been identified, the dataset can be filtered by those regions of the decision space and the sensitivity and conditional distribution analysis repeated. A new sensitivity profile may then emerge, because the importance of choices and uncertainties can vary across regions of the decision--uncertainty space. Importantly, zooming in is possible by fixing choices that are not contingent on uncertainty. Otherwise, the procedure filters out entire regions of uncertain futures instead of isolating the effects of the selected decision configurations. Finally, the process may also trigger model reformulation. If the observed performance profiles suggest missing design degrees of freedom, additional choices can be added and the multichoice modeling procedure repeated from the beginning.

\section{Cases}

This section presents three applications of multichoice modeling. We begin with a simple mortgage case that is accessible to readers regardless of background, and then proceed to two more specialized settings: a manufacturing line upgrade case and an M\&A integration strategy case.

In each case, we first perform a conventional analysis with a fixed decision architecture and then introduce multichoice modeling to demonstrate how decision support is enhanced when the decision configuration space is made explicit and explored jointly with uncertainty. The Discussion section synthesizes the insights that emerge from this shift in perspective and positions them relative to established decision-support approaches.

\subsection{Mortgage}

In Finland and many other countries, mortgage negotiations typically focus on reducing the interest-rate margin across offers from several banks, while the remaining terms are often treated as fixed. This case considers a typical mortgage offer, based on the authors' personal experience of mortgage negotiation in 2024. For a loan of 250,000 €, the negotiated margin was 0.6\%, implying a monthly payment of 1,199 €, as stated in the bank's offer (Table~\ref{tab:mortgage-details}).

\begin{table}[htbp]
\centering
\begin{minipage}[t]{0.42\textwidth}
\centering
\caption{Mortgage details}
\label{tab:mortgage-details}
\small
\begin{tabular}{lr}
\toprule
Item & Value \\
\midrule
Loan & €250,000 \\
Reference rate & 2.5\% \\
Margin & 0.6\% \\
Duration & 25 years \\
Monthly payment & €1,199 \\
\bottomrule
\end{tabular}
\end{minipage}
\hfill
\begin{minipage}[t]{0.48\textwidth}
\centering
\caption{Interest-rate parameters}
\label{tab:m-interest}
\small
\begin{tabular}{lr}
\toprule
Parameter & Value \\
\midrule
Normal mean level & 1\% \\
Mean reversion speed & 10\% \\
Volatility & 0.2\% \\
Crisis mean level & 6.5\% \\
Crisis duration & 5 years \\
Crisis occurrence & binary 50/50 \\
\bottomrule
\end{tabular}
\end{minipage}
\end{table}

This adds up to €359,572 paid over the 25-year loan period, corresponding to €109,572 in interest on top of the €250,000 principal. However, this figure is only a point estimate: the offer is based on a floating interest rate, so the actual interest expense depends on the future path of the reference rate. Estimating the range of possible interest expenses therefore requires modeling interest-rate uncertainty.

\subsubsection{Conventional analysis.} 

Although the mortgage choices are made at the outset, their consequences unfold through monthly payments over the loan period. To represent this time-dependent execution, the interest rate is modeled as a stochastic process rather than as a static range. Specifically, we use the mean-reverting Vasicek process \(r_{t+1} = r_t + a(b-r_t)\Delta t + \sigma \sqrt{\Delta t}\varepsilon_t,\) where $r_t$ denotes the interest rate at time $t$, $a$ is the speed of mean reversion, $b$ is the long-term mean level, $\sigma$ is the volatility, $\Delta t$ is the time step, and $\varepsilon_t \sim \mathcal{N}(0,1)$. For monthly simulations, the annual mean-reversion speed and volatility are adjusted by dividing them by 12 and $\sqrt{12}$, respectively.

\begin{figure}[htbp]
\centering
\includegraphics[width=\textwidth]{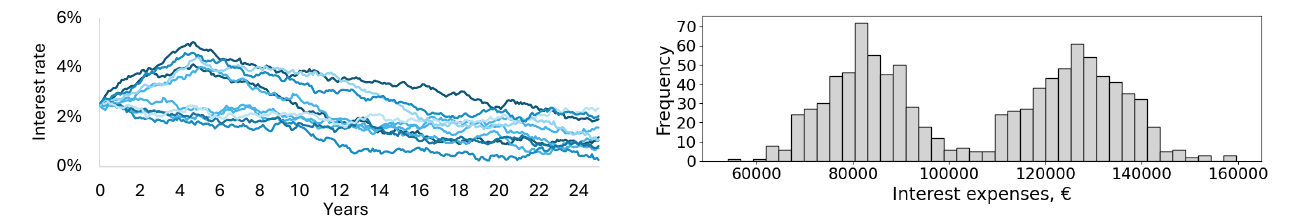}
\caption{Ten simulated paths of interest rate evolution under the mean-reverting process (left) and the resulting distribution of cumulative interest-rate expenses (right).}
\label{fig:m-hist}
\end{figure}

To reflect divergent expert estimates about future interest-rate paths, and to evaluate the resilience of alternative mortgage tactics, we model two scenarios for interest-rate evolution: ``normal times'' and ``crisis'', see Table \ref{tab:m-interest}. This bimodal stochastic process is visualized in Figure \ref{fig:m-hist}, left.

The resulting distribution of cumulative interest expenses has two clear peaks (Figure~\ref{fig:m-hist}, right), which is a direct consequence of the uncertainty scenario specification. The left peak corresponds to normal times, where cumulative interest expenses fall roughly between €50,000 and €100,000. The right peak reflects the simulated crisis scenario, in which expectedly total interest expenses can rise to approximately €160,000, or 64\% of the loan principal. Knowing the possible extent of future costs, however, does not yet tell us what to do about them.

\subsubsection{Multichoice modeling.} 

Once the decision is framed as a choice architecture rather than a fixed offer, several additional options become visible, even though they were not explicitly presented by the bank (Table~\ref{tab:mortgage-strategy-table}).

\begin{table}[htbp]
\small
\centering
\caption{Strategy table for mortgage choices}
\label{tab:mortgage-strategy-table}
\begin{tabular}{lll}
\hline
\textbf{Choice} & \textbf{Dependency} & \textbf{Effect in the model} \\
\hline
Rate type
& None
& Fixed 4.1\% / stochastic process \\

Interest rate cap
& Variable rate only
& 5\% cap at 0.05\% premium / no cap \\

Payment scheme
& Variable rate only
& Fixed payment / fixed term \\

Reference rate
& Variable rate only
& 3-month / 12-month Euribor \\
\hline
\end{tabular}
\end{table}

The first choice does not require simulation, since a fixed-rate mortgage produces a deterministic outcome that can be calculated directly using the model already developed. We therefore embed the remaining three choices into the model with interest-rate uncertainty. 

\begin{table}[htbp]
\centering
\begin{minipage}[t]{0.52\textwidth}
\centering
\small
\caption{Additional inputs for the multichoice model}
\label{tab:mortgage-multichoice-inputs}
\begin{tabular}{ll}
\hline
\textbf{Input} & \textbf{Value} \\
\hline
Interest rate cap, no/yes & Binary 50/50 \\
Cap level & 5\% \\
Cap fee & 0.05\% \\
Reference rate, 3-month/12-month & Binary 50/50 \\
3-month rate mean reversion speed & 20\% \\
3-month rate volatility & 0.3\% \\
Fixed payment/term & Binary 50/50 \\
\hline
\end{tabular}
\end{minipage}
\hfill
\begin{minipage}[t]{0.40\textwidth}
\centering
\small
\caption{Sensitivity indices}
\label{tab:mortgage-sensitivity-indices}
\begin{tabular}{lr}
\hline
\textbf{Variable} & \textbf{Index} \\
\hline
Crisis yes/no & 85.30\% \\
Fixed payment/term & 1.61\% \\
Reference rate, 3m/12m & 1.16\% \\
Interest rate cap, no/yes & 0.02\% \\
\hline
\end{tabular}
\end{minipage}
\end{table}

For this model structure, Equation~\ref{eq:sample-size} implies a benchmark sample size of \(1{,}600\) simulation runs. With 5,000 simulation runs, the case provides adequate sampling density for comparing full admissible configurations. The resulting simulation dataset records the binary input variables, namely the uncertainty scenario and the mortgage choices, together with the output variable, cumulative interest-rate expenses.

Variance-based sensitivity analysis identifies the uncertainty scenario as the dominant driver of interest-rate expenses. This is expected, given the intentionally strong contrast between the normal-times and crisis scenarios. More importantly for decision support, two mortgage choices receive non-zero sensitivity indices, as shown in Table~\ref{tab:mortgage-sensitivity-indices}.

\begin{figure}[htbp]
\centering
\includegraphics[width=\textwidth]{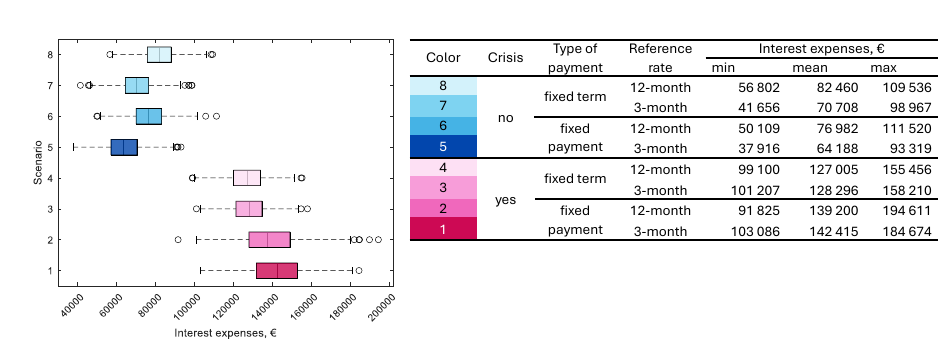}
\caption{Conditional distributions produced by the combination of the dominant uncertainty variable and two mortgage choices.}
\label{fig:m-dec}
\end{figure}

Figure~\ref{fig:m-dec} depicts the conditional distributions produced by combinations of payment schemes and reference-rate maturities under the normal-times (blue) and crisis (red) scenarios. Although the corresponding sensitivity indices are low, the choices remain economically meaningful: even under stable conditions, the difference between the best and worst strategies is approximately €20,000.


The relative performance of the payment schemes reverses across scenarios. Under crisis conditions, the fixed-payment scheme generates higher cumulative interest expenses than the fixed-term plan because rising rates shift a larger share of payments toward interest, extend the loan term, and increase total borrowing costs. Under normal conditions, declining rates shorten the loan under the fixed-payment scheme, making it less costly than the fixed-term alternative. Thus, fixed payment performs better in stable environments, whereas fixed term is more robust under adverse interest-rate paths. 

Nonlinear effects are also observed for reference-rate maturity. Under crisis conditions, reference-rate maturity has almost no effect. Under normal conditions, however, the 3-month rate yields consistently lower interest costs than the 12-month rate. This follows from its faster mean reversion, despite its higher volatility. Historical Euribor data are consistent with this pattern: in the observed data, the 3-month rate has been lower than the 12-month rate in more than 95\% of observations. The cost advantage of the shorter maturity is therefore empirically supported.

The interest-rate cap, omitted from the decomposition in Figure~\ref{fig:m-dec}, has near-zero importance (Table~\ref{tab:mortgage-sensitivity-indices}). Because the offered 5\% cap is rarely triggered in the simulated interest-rate paths (Figure~\ref{fig:m-hist}, left), the added margin premium provides little protection while increasing expected costs.

Finally, a fixed-rate mortgage at 4.1\% results in total interest expenses of approximately €150,000, exceeding most simulated outcomes, including many crisis cases. It is therefore not competitive in this setting.

Overall, the fixed-payment and 3-month reference-rate combination performs best under normal conditions but becomes one of the worst options in crisis. In contrast, the fixed-term and 3-month reference-rate configuration performs well in both uncertainty scenarios. It is therefore the most robust choice across the modeled futures, unlike the default offer based on fixed payments and a 12-month reference rate.

The shift from conventional mortgage evaluation to multichoice modeling not only makes the available choices explicit but also crucially changes the recommendation. A conventional analysis asks how risky the offered mortgage is under possible interest-rate paths. The multichoice model asks which contract configuration is preferred once payment scheme, reference-rate maturity, and protection mechanisms are treated as explicit choices. The resulting recommendation is a reconfiguration of the decision object itself.

\subsection{Manufacturing process upgrade}

This case is based on a project with Hyperion Robotics, an innovative Finnish construction company. The project was conducted at the end of 2023 and documented in detail by \cite{walzer2026scale}. In this paper, we present a reduced model for ease of exposition while preserving the relevant complexity of the decision space, the actual progression of the project, and the insights derived at each stage. Some details are anonymized in accordance with the company's requirements.

The company already had a deterministic model for computing the unit cost of its 3D-printed construction element. The model links production parameters to the associated cost components, with unit cost defined as the ratio of total daily cost to daily output. The production process consists of three stages: preparation, printing, and post-processing. Preparation and post-processing activities together require 5.6 hours per operational day. During the printing stage, the robotic system operates at a nominal material deposition rate of 2 tons per hour, corresponding to 0.83 m$^3$/h for a concrete density of 2.4 t/m$^3$. The intermittent installation of steel reinforcement elements reduces effective productivity to 60\% of the nominal printing rate. 

With these assumptions (Table~\ref{tab:unit-cost-inputs}), the model returns a total daily cost of €4,389. Given a daily output of 0.96 units, this corresponds to a current unit cost of €4,572 per unit.

\begin{table}[htbp]
\centering
\begin{minipage}[t]{0.48\textwidth}
\vspace{0pt}
\centering
\small
\caption{Input variables in the unit cost model}
\label{tab:unit-cost-inputs}
\begin{tabular}{ll}
\hline
\textbf{Input variable} & \textbf{Value} \\
\hline
\multicolumn{2}{l}{\textit{Production}} \\
Non-printing time, h/day & 5.6 \\
Volume, m$^{3}$/unit & 1.25 \\
Printing speed, ton/h & 2 \\
Robotic system productivity & 60\% \\
\hline
\multicolumn{2}{l}{\textit{Costs}} \\
Workers, persons & 6 \\
Salary, €/h & 22 \\
Buffer/waste, \% concrete use & 100\% \\
Concrete price, €/ton & 441 \\
Steel reinforcement, \% volume & 1\% \\
Steel reinforcement price, €/kg & 2 \\
Self-compacting concrete, m$^{3}$/unit & 0.63 \\
Self-compacting concrete price, €/m$^{3}$ & 1000 \\
\hline
\end{tabular}
\end{minipage}
\hfill
\begin{minipage}[t]{0.48\textwidth}
\vspace{0pt}
\centering
\small
\caption{Improvement potential in input variables}
\label{tab:input-variable-variation}
\resizebox{\linewidth}{!}{%
\begin{tabular}{lll}
\hline
\textbf{Variable} & \textbf{Current} & \textbf{Better} \\
\hline
Concrete price, €/ton & 441 & 223 \\
Buffer/waste & 100\% & 50\% \\
Self-compacting concrete price, €/m$^{3}$ & 1000 & 300 \\
Preparation time, h/day & 5.6 & 2 \\
Printing speed, ton/h & 2 & 4 \\
Robotic system productivity & 60\% & 100\% \\
\hline
\end{tabular}%
}
\end{minipage}
\end{table}

\subsubsection{Conventional analysis.} 
\begin{figure}[htbp]
\centering
\includegraphics[width=\textwidth]{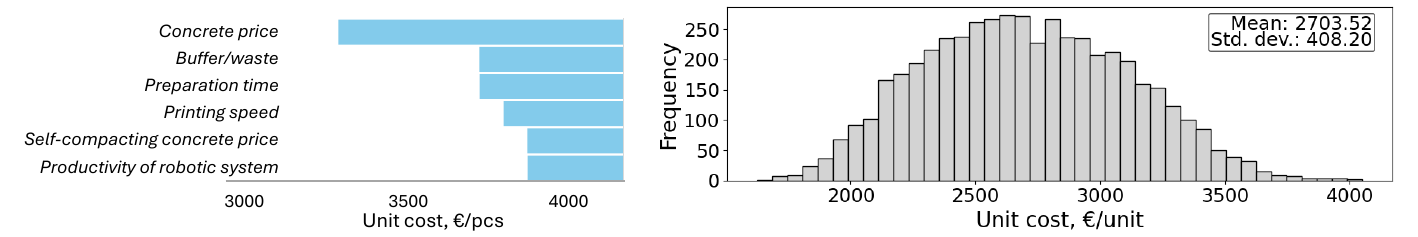}
\caption{One-at-a-time sensitivity analysis (left) and Monte Carlo simulation (right) results.}
\label{fig:hyp-hist}
\end{figure}

The initial intuition was to map which input variables in the existing model could be improved, and by how much, and then to perform two analyses: (i) a one-at-a-time sensitivity analysis to identify which variables had the strongest effect on unit cost, and (ii) a Monte Carlo simulation to examine their joint effect. The company provided estimates of the feasible improvement ranges for several model inputs, as shown in Table~\ref{tab:input-variable-variation}. These values were first used in a basic sensitivity analysis, varying one input at a time. For the Monte Carlo simulation, uniform distributions between the current and improved values were assumed, since these parameter ranges represented possible improvements rather than probabilistic uncertainty. The results are presented in Figure~\ref{fig:hyp-hist}.

The results confirmed the company's intuition that the concrete price was the most influential variable. More unexpectedly, the analysis showed that the combined improvement of the selected inputs could reduce the unit cost by roughly half. 

However, these results were not sufficient to guide an investment decision. Each variable in the one-at-a-time sensitivity analysis was, in practice, the result of one or more possible upgrades, and these upgrades could affect several variables simultaneously.

\subsubsection{Multichoice modeling.}

To support the decision on which upgrades to implement, the upgrades themselves needed to be incorporated into the model. Each upgrade affects specific subsets of model variables in predefined ways. For instance, material cost is no longer represented by a single parameter; instead, the model includes three alternative material choices, each with a distinct unit price. Their availability depends on the hardware configuration, since certain materials can only be used if specific hardware upgrades are selected. The production system is also subject to a space-related capacity constraint, which currently limits output to two units per day. This capacity can be increased through additional hardware upgrades that modify the corresponding capacity parameter. Finally, each upgrade has its own investment cost, so improvements in unit cost must be evaluated against the required capital expenditure.

Listing the upgrades together with their effects on model variables and their interdependencies naturally produced a strategy table, shown in Table~\ref{tab:upgrade-options-effects}. 

\begin{table}[htbp]
\centering
\caption{Upgrade options, effects, and computed sensitivity indices}
\label{tab:upgrade-options-effects}
\small
\begin{tabularx}{\textwidth}{p{0.22\textwidth}rXr}
\toprule
\textbf{Area/upgrade} & \textbf{Cost, K€} & \textbf{Effect} & \textbf{Sensitivity index} \\
\midrule
\multicolumn{3}{l}{\textit{Material}} & 3\% \\
Concrete 2 & 40 & Price 301 €/ton; available with Mixer upgrade. & \\
Concrete 3 & 40 & Price 223 €/ton; available with Mixer and Pump upgrades. & \\
\midrule
\multicolumn{3}{l}{\textit{Hardware}} & \\
Mixer & 90 & Enables self-compacting concrete up to 300; enables Concrete 2. & 20\% \\
Printing head 1 & 50 & Reduces labor by one worker. & 0\% \\
Printing head 2 & 100 & Reduces buffer/waste to 10\%; changes space constraint to 4 units per day; reduces non-printing time by 0.5 h. & 22\% \\
Robot & 300 & Changes space constraint to 6 units per day. & 1\% \\
Pump & 100 & Reduces non-printing time by 1.5 h; reduces buffer/waste to 10\%; reduces labor by two workers; increases productivity to 100\%; enables Concrete 3 if Mixer upgrade is selected. & 49\% \\
Steel reinforcement & 150 & Reduces non-printing time by 0.5 h; reduces labor by one worker; changes productivity to 80\%. & 2\% \\
\midrule
\multicolumn{3}{l}{\textit{Software}} & \\
Print speed software & 25 & Increases printing speed, ton/h, by a factor of 1.5. & 1\% \\
Software library & 200 & Reduces non-printing time by 1.6 h. & 1\% \\
\bottomrule
\end{tabularx}
\end{table}

To integrate investment costs into the unit cost calculation, the former are converted into daily annuities using a discount rate of 5\% and a time horizon of three years. All technology upgrades are modeled as binary choices. The company first wanted to examine the upgrade choice architecture alone, while keeping the workflow simulation-ready so that uncertainty sources could be introduced in subsequent analyses. 

Equation~\ref{eq:sample-size} implies a benchmark sample size of \(44{,}800\) simulation runs for such a model. However, this case is not intended to support comparison of all full admissible configurations. Its purpose is to identify the first moves that most strongly shape performance. We therefore use an approximately one-tenth of the benchmark size, \(5{,}000\) runs. A robustness check with a larger sample led to the same conclusion.

Variance-based sensitivity analysis identified \textit{Pump upgrade}, \textit{Printing head upgrade 2} and \textit{Mixer upgrade} as the most influential, Table \ref{tab:upgrade-options-effects}, last column. All other choices gained negligible indices. Interestingly, \textit{Material} yielded a small sensitivity index, even though it directly defines the concrete price, which was flagged as the most influential model input in the one-at-a-time analysis of a non-multichoice model.

\begin{figure}[htbp]
\centering
\includegraphics[width=\textwidth]{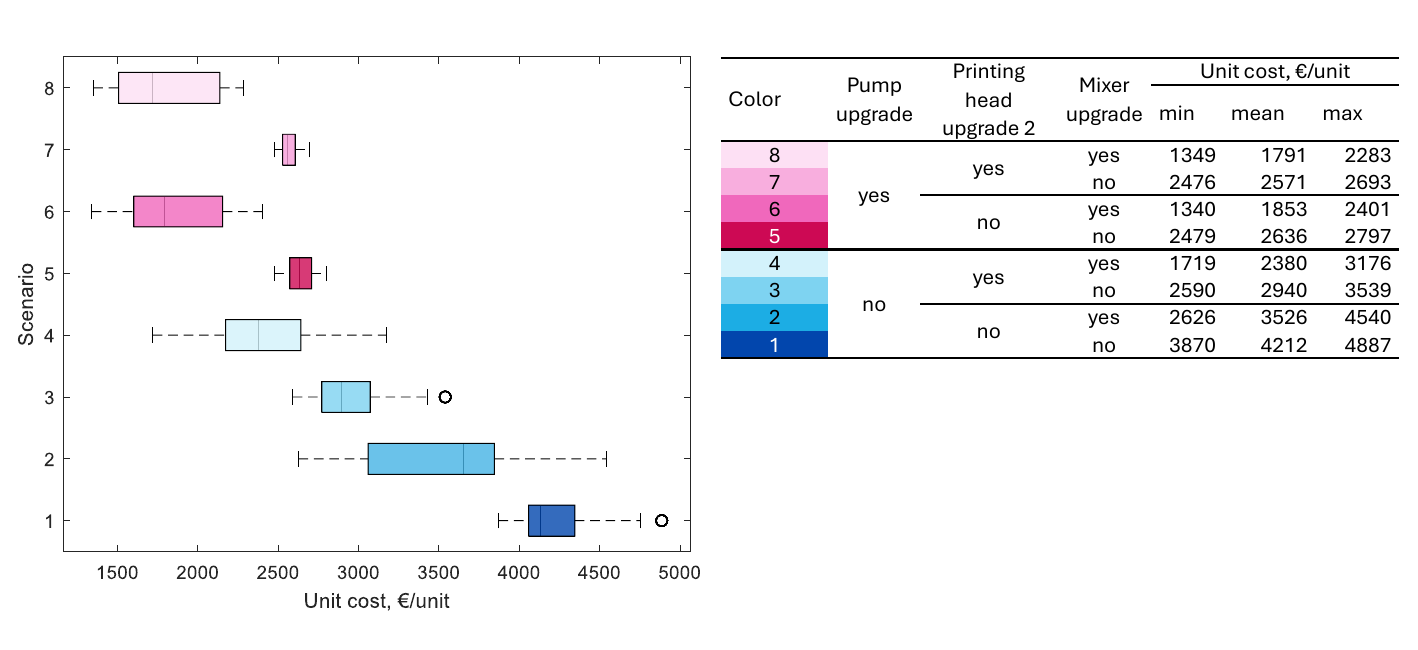}
\caption{Conditional distributions of unit cost across configurations defined by the three most influential choices.}
\label{fig:hyp-dec1}
\end{figure}

Figure~\ref{fig:hyp-dec1} decomposes the simulation results into conditional distributions defined by these three upgrades. The analysis of upgrade combinations confirms that the configuration with all three key upgrades (\textit{Pump upgrade}, \textit{Printing head upgrade 2}, and \textit{Mixer upgrade}) yields the lowest observed unit cost, while the configuration with none of them yields the highest. Between these extremes, however, the effects are strongly nonlinear.

The \textit{Printing head upgrade 2} substantially reduces unit cost when the \textit{Pump upgrade} is not implemented. In this case, the corresponding boxplots, cases 3 and 4, are shifted markedly to the left relative to the baseline without the \textit{Printing head upgrade 2}, as in cases 1 and 2. When the \textit{Pump upgrade} is already in place, however, the marginal effect of the \textit{Printing head upgrade 2} becomes negligible: the distributions with and without it, cases 5 versus 7 and 6 versus 8, are nearly indistinguishable. This indicates a strong overlap in the effect mechanisms of the \textit{Pump upgrade} and \textit{Printing head upgrade 2}. Moreover, a relatively low unit cost can be achieved without relaxing the space constraint, corresponding to production remaining below the two-units-per-day limit.

The \textit{Mixer upgrade} has a substantial effect even when the \textit{Pump upgrade} is already implemented. Without the \textit{Mixer upgrade}, the unit cost distribution is narrower and remains at a higher level, as shown by cases 5 and 7. Activating the \textit{Mixer upgrade} widens the distribution and shifts it downward. This occurs because the \textit{Mixer upgrade} enables the use of \textit{Concrete 2}; when combined with the \textit{Pump upgrade}, it also enables \textit{Concrete 3}. Because these materials have lower prices, they widen the conditional unit cost distributions by extending them towards lower values.

\begin{figure}[htbp]
\centering
\includegraphics[width=0.8\textwidth]{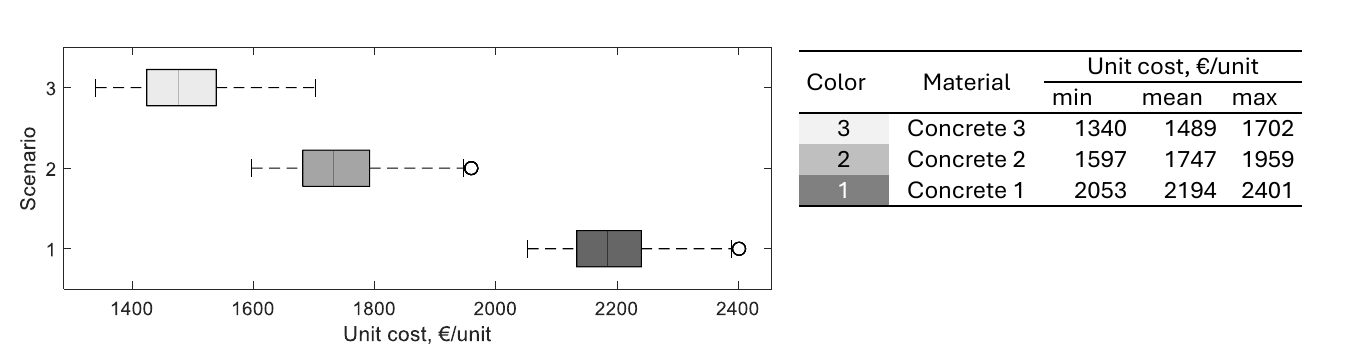}
\caption{Conditional distributions for material choices within the subset of configurations with Pump and Mixer upgrades.}
\label{fig:hyp-dec2}
\end{figure}

In the variance-based sensitivity analysis across all simulated scenarios, the \textit{Material} variable has a relatively low sensitivity index of 3\%, despite its evident importance in certain configurations. This pattern is consistent with highly nonlinear models, where a factor may have strong influence within a restricted subset of the design space but a diluted average impact across all combinations. When the analysis is conditioned on scenarios in which both the \textit{Pump upgrade} and \textit{Mixer upgrade} are active, cases 6 and 8, material choice becomes the dominant driver of unit cost with a sensitivity index of 92\%. In this subset, \textit{Concrete 3} yields the lowest unit costs, with values below €1,700 per unit, Figure \ref{fig:hyp-dec2}.

The analysis indicates that the largest reductions in unit cost are obtained by implementing the \textit{Pump upgrade}, followed by the \textit{Mixer upgrade}, and then adopting \textit{Concrete 3}. This sequence targets the largest unit cost reductions while keeping the implementation path focused on a small number of simultaneous improvements. Although \textit{Printing head upgrade 2} has a high sensitivity index, its effect is material only in configurations where the \textit{Pump upgrade} is absent. Given its limited marginal contribution under the recommended upgrade path, it is therefore not included. Conversely, \textit{Material} has a low sensitivity index in the full simulated design space, but becomes the primary driver of unit cost within the subset of scenarios where both the \textit{Pump upgrade} and \textit{Mixer upgrade} are active. Therefore, \textit{Concrete 3} is included as the final step once these enabling upgrades are in place.

This case presents two different models of the same underlying production system. The first follows a conventional modeling practice: it computes unit cost as a function of operational parameters and supports basic model analysis, but it does not directly indicate which actions the company should take. The second is a multichoice model: it encodes the behavior of the unit cost model as a function of available upgrade choices, including their dependencies and enabling conditions. As a result, this level of analysis reveals not only which variables matter, but which choices need to be undertaken, in what sequence, and under which configurations. Although both models compute unit cost, the emphasis of the decision support outcome is entirely different. The conventional model asks what would happen if selected input parameters changed. The multichoice model asks what the company can actually do to make those changes happen. This distinction is central because actionable decision support requires moving from sensitivity to variables toward analysis of the choice architecture. 

\subsection{M\&A strategy}
\label{sec:mna}

This synthetic case was constructed based on consultations with two independent senior M\&A experts based in Finland. In practice, acquisitions are rarely analyzed as a space of alternative choices. Instead, decision makers typically formulate a single acquisition strategy, define an integration approach, and evaluate the transaction through identified synergies. A discounted cash flow valuation is then used to compare the estimated value of the target with the proposed purchase price and support the final decision.

In the modeled case, an equipment manufacturing company considers acquiring a target firm operating in the equipment service and refurbishment market. The acquisition is motivated by two potential synergies. First, the buyer expects market synergies through cross-selling opportunities between the customer bases of the two businesses. Second, integration may create operational flexibility, allowing capacity to be shifted between manufacturing and service activities as their relative attractiveness changes over time. The baseline characteristics of the businesses are summarized in Table \ref{tab:mna-baseline}.

\begin{table}[htbp]
\centering
\begin{minipage}[t]{0.45\textwidth}
\vspace{0pt}
\centering
\small
\caption{Baseline assumptions of the M\&A case}
\label{tab:mna-baseline}
\begin{tabular}{lr}
\toprule
\textbf{Parameter} & \textbf{Value} \\
\midrule
Deal price, M€ & 80 \\
Existing business EBITDA, M€/year & 25 \\
Service EBITDA, M€/year & 10 \\
Equipment capacity, units & 80 \\
Service capacity, units & 20 \\
\bottomrule
\end{tabular}
\end{minipage}
\hfill
\begin{minipage}[t]{0.50\textwidth}
\vspace{0pt}
\centering
\small
\caption{Uncertainty assumptions in the M\&A case}
\label{tab:mna-uncertainties}
\begin{tabular}{lr}
\toprule
\textbf{Parameter} & \textbf{Value} \\
\midrule
Integration success & \(U(0.4,1.0)\) \\
Market synergy potential, M€/year & \(U(0,10)\) \\
Equipment market attractiveness & Random walk \\
\quad Initial value & 1.0 \\
\quad Volatility & 0.20 \\
Service market attractiveness & Random walk \\
\quad Initial value & 1.0 \\
\quad Volatility & 0.10 \\
\bottomrule
\end{tabular}
\end{minipage}
\end{table}

\subsubsection{Conventional analysis.}

\begin{figure}[htbp]
    \centering
    \includegraphics[width=\textwidth]{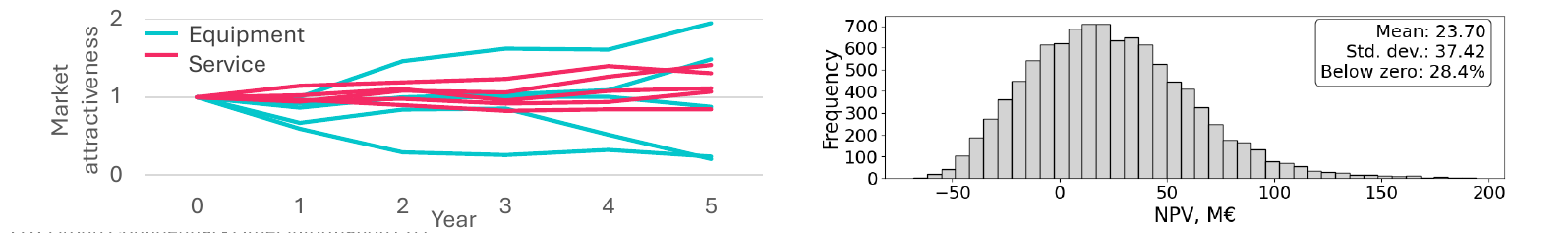}
    \caption{Five sample realizations of the stochastic equipment and service market attractiveness processes (left) and distribution of simulated net present value (NPV) outcomes for the M\&A case (right).}
    \label{fig:mna-unc}
\end{figure}

To incorporate the value of production-mix flexibility, the attractiveness of the equipment and service markets is modeled stochastically through random walk. This allows the two markets to develop distinctly over time, creating opportunities for capacity reallocation. The equipment market is modeled as more volatile compared to the service market, as illustrated in Figure \ref{fig:mna-unc} (left). Other uncertainties include market synergy potential and integration success. Numeric assumptions are summarized in Table \ref{tab:mna-uncertainties}.

The value of operational flexibility is captured through the possibility of reallocating production capacity between equipment manufacturing and service activities. The maximum share of capacity that can be reallocated is determined by integration success, $F = 0.5 \cdot \textit{integration success}$, reflecting the idea that successful integration improves the organization's ability to coordinate resources across business units.

In each year $t$, profitability per unit of capacity is calculated separately for equipment manufacturing and service activities as $\pi_{i,t}=A_{i,t}EBITDA_i/C_i$, where $A_{i,t}$ is market attractiveness, $EBITDA_i$ is the baseline EBITDA of business $i$, and $C_i$ is its production capacity.

The annual capacity reallocation is then determined as
\begin{equation}
\Delta C_t =
F
\frac{\pi_{E,t}-\pi_{S,t}}
{\pi_{E,t}+\pi_{S,t}}.
\label{eq:capacity-shift}
\end{equation}

Positive values of $\Delta C_t$ indicate capacity reallocation towards equipment manufacturing, whereas negative values indicate reallocation in favor of service activities.

Enabling operational flexibility requires a one-time investment of €10M. Annual flexibility costs are assumed equal to $10(1-\textit{integration success})$ M€, representing coordination inefficiencies and organizational friction associated with operating a flexible production system.

Integration costs are modeled as $5+30(1-\textit{integration success})$ M€, reflecting the greater effort and disruption associated with less successful integrations. Market development incurs a deterministic cost of €2M per year.

The profitability of the acquisition is evaluated over a five-year horizon using a discounted cash flow model with a 5\% discount rate. A terminal value equal to four times the final-year EBITDA is added at the end of the planning horizon.

These modeling assumptions translate into a distribution of M\&A value concentrated between $-50$M€ and $150$M€ (Figure \ref{fig:mna-unc}, right). While the overall outlook is reasonably favorable, with only about one-third of outcomes resulting in negative values, the analysis remains limited to evaluating a predefined acquisition strategy under uncertainty. It cannot reveal whether the strategy itself is appropriate, which strategic choices drive performance, and whether the downside risk can be mitigated through alternative acquisition, integration, or market development approaches.

\subsubsection{Multichoice modeling.}

The uncertainty analysis above assumes that the acquisition strategy is fixed. However, acquisition performance depends not only on uncertain future conditions but also on how the deal is structured and managed. To make these strategic degrees of freedom explicit, the model is reformulated as a multichoice model. The acquisition is represented as a set of interacting choices concerning acquisition staging, information use, integration, operational flexibility, and market development, Table \ref{tab:mna-strategy-table}.

\begin{table}[htbp]
\centering
\caption{Choice structure of the M\&A case}
\label{tab:mna-strategy-table}
\small
\begin{tabular}{p{2.8cm}p{3.0cm}p{5.0cm}p{4.6cm}}
\toprule
\textbf{Choices} &
\textbf{Dependencies} &
\textbf{Effects} &
\textbf{Numeric assumptions} \\
\midrule

\textbf{Staged} \newline \textbf{acquisition} \newline
$\bullet$ No \newline
$\bullet$ Yes
&
None
&
Delays full commitment and enables learning about integration success before the remaining stake is acquired.
&
First-stage price is €35M. Learned integration success takes values low (40--60\%), medium (60--80\%), or high (80--100\%).
\\
\addlinespace

\textbf{Buying trigger} \newline
$\bullet$ Low \newline
$\bullet$ Medium \newline
$\bullet$ High
&
Relevant only if staged acquisition = Yes
&
Determines the threshold of learned integration success required to proceed with full acquisition.
&
Directly tied to the levels of learned integration success. If learned integration success $\geq$ trigger, then buy the remaining stake and integrate fully. Second-stage price is €55M with a one-year delay.
\\
\addlinespace

\textbf{Integration} \newline
$\bullet$ Partial \newline
$\bullet$ Full
&
If buy remaining stake = Yes, only Full integration is possible
&
Full integration increases market synergy realization and enables operational flexibility, but incurs higher integration costs.
&
Full integration cost (M€): $5 + 30(1-\text{integration success})$.
\\
\addlinespace

\textbf{Operational} \newline
\textbf{flexibility} \newline
$\bullet$ No \newline
$\bullet$ Yes
&
Available only if integration = Full
&
Allows capacity reallocation between equipment manufacturing and service activities in response to changing market conditions; incurs recurring flexibility costs dependent on integration success.
&
Investment cost is €10M. Recurring costs are $10(1-\text{integration success})$ M€/year.
\\
\addlinespace

\textbf{Market} \newline
\textbf{development} \newline
$\bullet$ Passive \newline
$\bullet$ Active
&
None
&
Determines the extent to which potential market synergies are pursued through cross-selling and market expansion; active development incurs additional costs.
&
Cost of active market development is €2M/year.
\\

\bottomrule
\end{tabular}
\end{table}

Some of these choices have been previously discussed in the real options literature, where acquisitions may be viewed as options to expand, defer, stage, or abandon investments \citep{alvarez2006takeover,CollanKinnunen2009AcquisitionStrategy}. However, these options are typically analyzed individually or within a predefined acquisition strategy. Multichoice modeling represents available real options jointly as interacting strategic choices whose performance can be explored jointly under uncertainty.

Only operational flexibility fits the classical real-options setup directly: switching between manufacturing and service activities is exercised in response to evolving uncertainty. Investment-stage real options, such as staging and learning, are typically modeled by optimizing over future uncertain payoffs. In multichoice modeling, however, such optimization would conflict with the premise of a rugged decision landscape, where the value of an option may depend on the surrounding configuration of choices. We therefore do not optimize the staged acquisition option directly. Instead, we reformulate it as a choice over decision triggers: if the acquisition is staged, the remaining stake is acquired when integration success is known to have reached a low, medium, or high threshold.

To facilitate interpretation, the stochastic market trajectories are aggregated into a categorical variable, \textit{Market}. The variable takes values 0 (equipment-dominated), 1 (balanced), and 2 (service-dominated), based on the difference between the cumulative equipment and service market-attractiveness indices. Differences within $\pm 1$ are classified as balanced, while larger deviations are assigned to the market with the higher cumulative attractiveness.

In addition to acquisition NPV, the value of operational flexibility is computed as the incremental cash flows generated by reallocating production capacity between equipment manufacturing and service activities. An ample simulation design of 100,000 runs was used.

\begin{figure}[htbp]
\centering
\includegraphics[width=\textwidth]{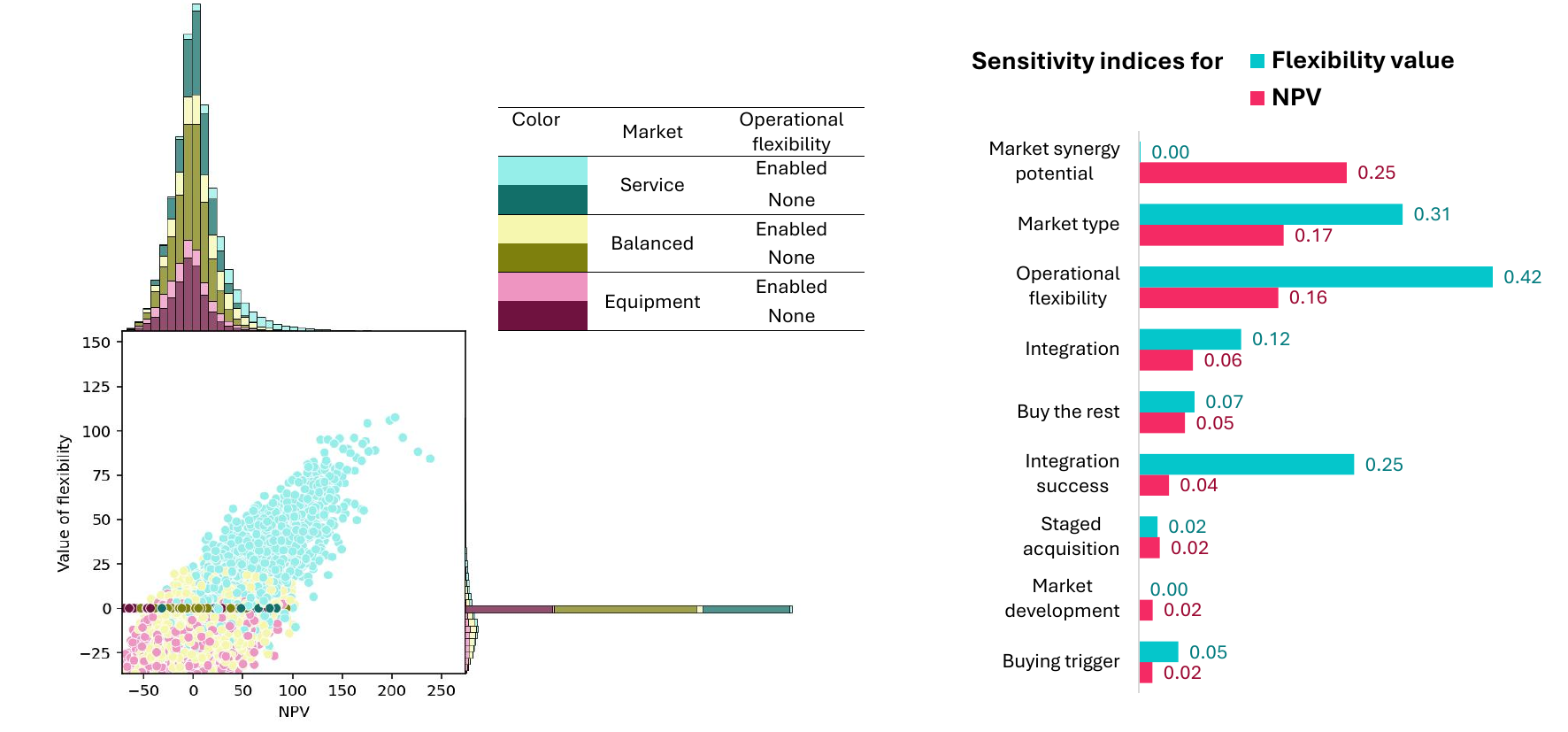}
\caption{Joint analysis of M\&A profitability and operational flexibility value: simulated outcomes (left) and sensitivity indices (right).}
\label{fig:mna-two-output}
\end{figure}

Figure \ref{fig:mna-two-output} compares the distributions and sensitivity profiles of both performance measures. Conditional distributions are constructed out of the two most influential variables for both output measures: uncertain market and the choice to enable operational flexibility. 

Figure \ref{fig:mna-two-output} reveals that operational flexibility creates value primarily when service markets outperform equipment markets, which also generate the most profitable acquisition outcomes. At the same time, many simulation outcomes remain profitable even without operational flexibility, provided that service markets dominate. Balanced and equipment-dominated markets can also result in positive acquisition value, but they are associated with substantially greater downside risk.

\begin{figure}[htbp]
\centering
\includegraphics[width=0.5\textwidth]{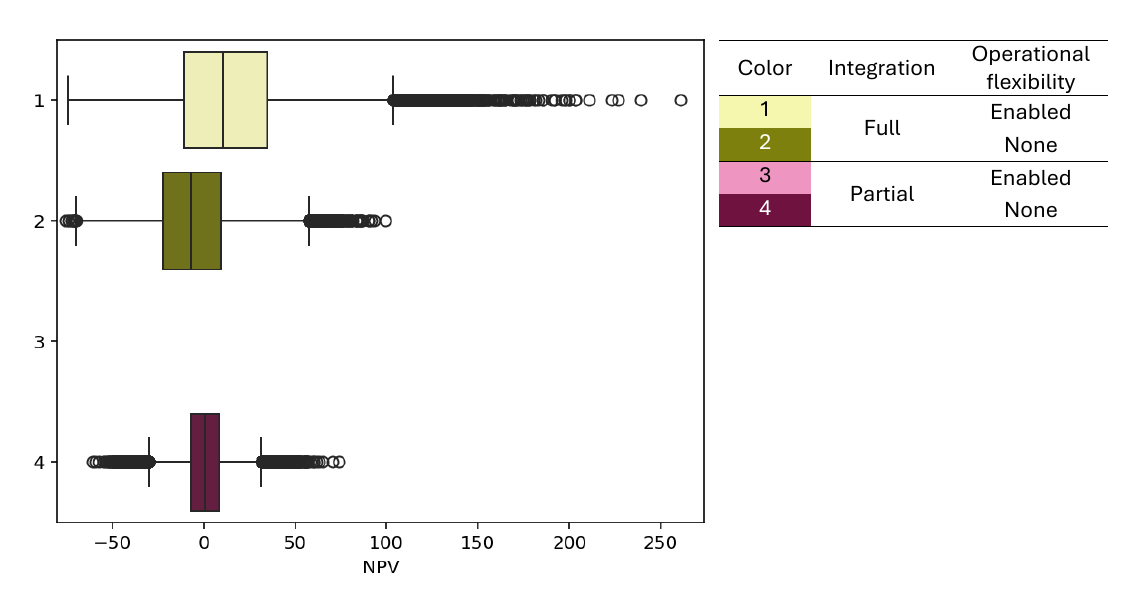}
\caption{Conditional distributions of acquisition NPV by integration and operational flexibility choices.}
\label{fig:mna-int-flex}
\end{figure}

The second most important choice affecting acquisition NPV, after operational flexibility, is the integration approach. Conditioning NPV jointly on integration and operational flexibility choices (Figure \ref{fig:mna-int-flex}) further supports the conclusion that operational flexibility is the primary source of value creation. Full integration without operational flexibility erodes acquisition value, suggesting that integration is not a value creator in itself but an enabling mechanism for operational flexibility.

Since market conditions and market synergy potential are the dominant sources of uncertainty, the choice configurations created by operational flexibility and integration were further examined across different states of uncertainty. The observed pattern remained stable, suggesting that the conclusion is robust to variations in market conditions and synergy realization.

\begin{figure}[htbp]
\centering
\includegraphics[width=\textwidth]{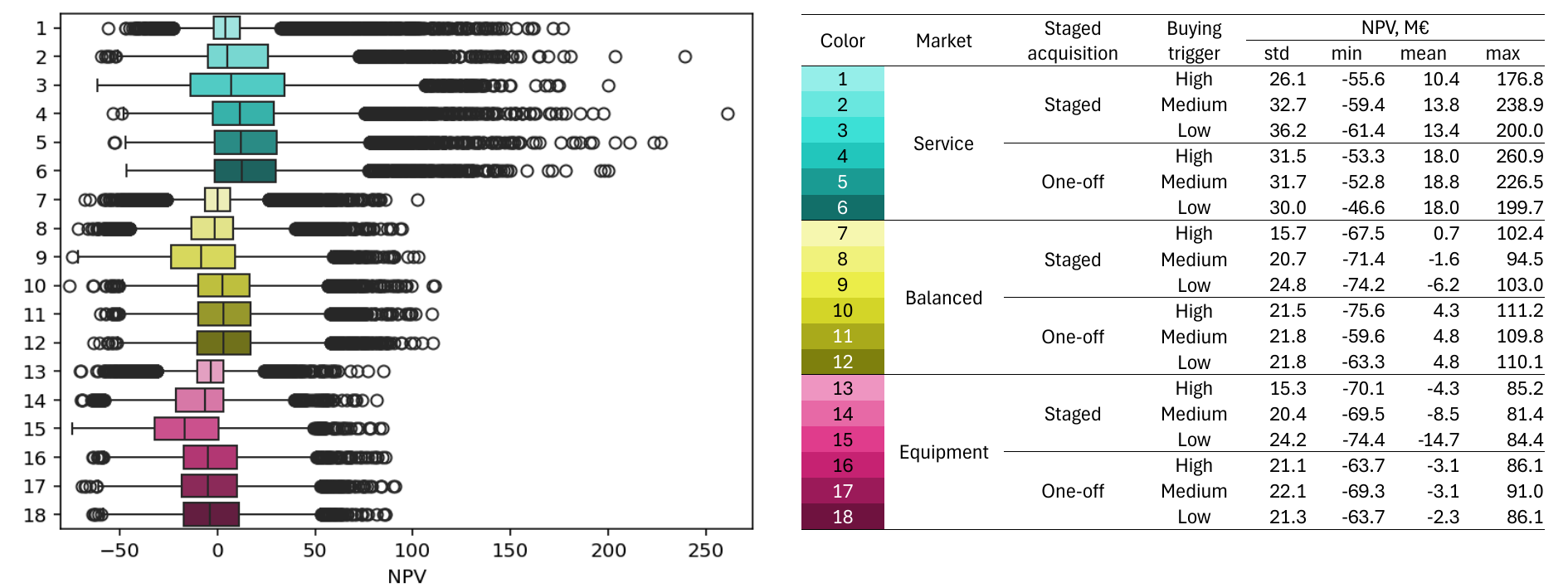}
\caption{Conditional distributions of acquisition NPV by market type, acquisition staging, and buying trigger choices.}
\label{fig:mna-market-staged-trigger}
\end{figure}

Another important set of choices, although not highlighted by the sensitivity indices, concerns acquisition staging and the trigger for acquiring the remaining stake. Unlike the previously discussed choices, the performance of these configurations changes across the types of market uncertainty. Therefore, they are analyzed conditional on market type, as shown in Figure \ref{fig:mna-market-staged-trigger}.

Figure \ref{fig:mna-market-staged-trigger} first separates the results into three market types, highlighted by the color groups. Within each market type, the lower three boxes correspond to acquiring the entire target company immediately. In these configurations, the trigger choice is irrelevant because no learning takes place before the acquisition; consequently, the three boxes are nearly identical, with only minor differences attributable to the simulation noise. The top three boxes correspond to staged acquisition. Here, each box represents a different trigger level that determines whether the remaining stake is acquired after observing the realized integration success.
 
In service-dominated markets, lower acquisition triggers generate the highest expected value (as seen in the comparison of boxes 3 to 1) by preserving access to the substantial benefits of operational flexibility. In contrast, balanced and equipment-dominated markets favor more conservative triggers (as seen in the comparison of boxes 7 to 9 and 13 to 15), which reduce exposure to unfavorable acquisitions. Consequently, no globally optimal trigger exists. A trigger that improves performance under one market type can degrade it under another, illustrating why optimizing individual real options in isolation may be misleading in a rugged decision-uncertainty space.

Independent of market type, higher trigger levels reduce the volatility of acquisition outcomes by screening out a larger share of unfavorable acquisitions. Nevertheless, all trigger levels produce lower expected value than immediate acquisition. Thus, despite the availability of staging and learning options, the preferred strategy under the considered uncertainties and choice configurations is to acquire the target company immediately.

\begin{figure}[htbp]
\centering
\includegraphics[width=\textwidth]{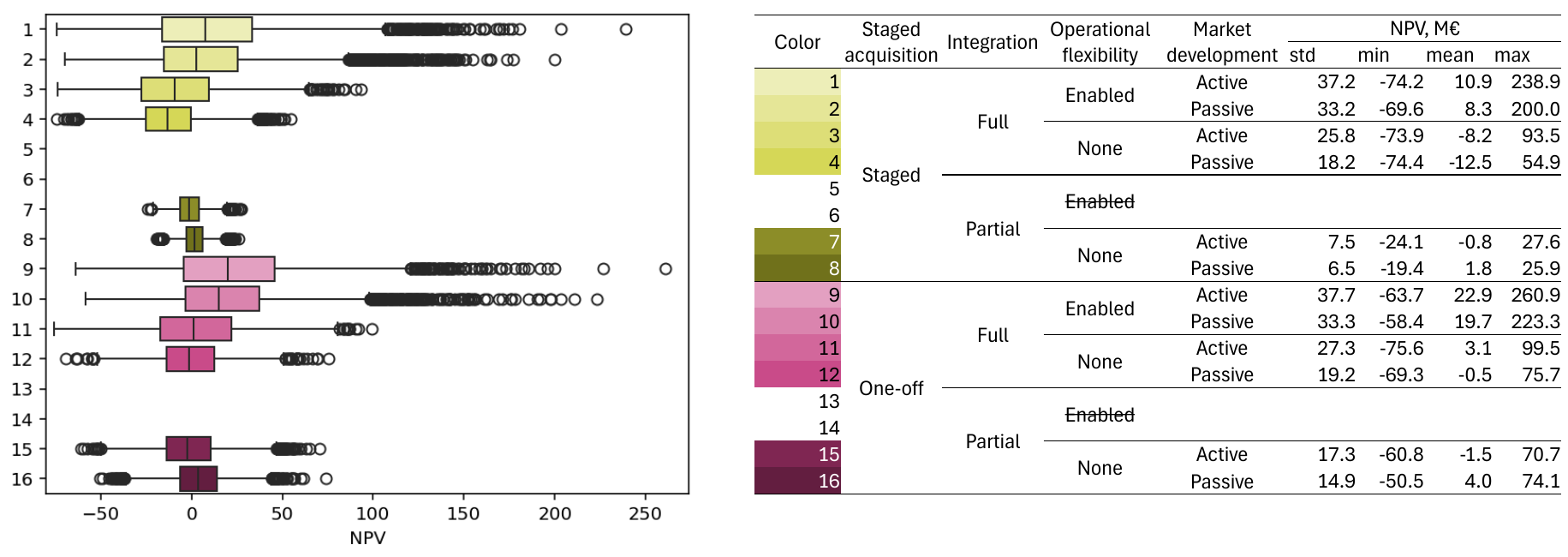}
\caption{Conditional distributions of acquisition NPV by acquisition staging, integration, operational flexibility, and market development choices.}
\label{fig:mna-market-dev}
\end{figure}

The final choice concerns market development. Having established the preferred acquisition, integration, and operational flexibility configurations, the analysis now conditions on these choices and examines whether active market development contributes to value (Figure \ref{fig:mna-market-dev}).

Within the preferred acquisition configurations, active market development improves both the expected value and the upside potential of the deal (boxes 1 and 2 in Figure \ref{fig:mna-market-dev}). As expected, however, the additional investment also increases downside exposure by introducing recurring development costs. Interestingly, the value of market development depends on the surrounding configuration of other choices. Under partial integration and without operational flexibility, passive market development becomes preferable (as seen in the comparison of boxes 7 and 8, as well as 15 and 16). Thus, even this seemingly straightforward choice does not have a universally preferred alternative; its attractiveness depends on the broader configuration of strategic choices.

The overall strategy recommended by the multichoice analysis is to acquire the target company immediately, perform full integration, invest in operational flexibility, and pursue active market development. Notably, this is the same strategy that was evaluated in the preceding uncertainty-only analysis. However, the multichoice analysis provides a much stronger justification for this recommendation. Rather than merely quantifying the uncertainty surrounding a predefined strategy, it systematically explores alternative acquisition architectures and demonstrates why competing configurations are inferior.

This case illustrates three important points. First, real options such as acquisition staging, learning, and operational flexibility can be naturally represented within a multichoice modeling framework. Second, optimizing individual options in isolation may be misleading on a rugged decision landscape, as their attractiveness depends on the surrounding configuration of choices and uncertainties. Third, both strategic choices and real options can either create or destroy value depending on how they interact with one another and with the evolution of uncertain future conditions.

\subsection{Discussion}


The three cases highlight how multichoice modeling extends established decision-support perspectives by shifting the focus from evaluating predefined strategies to exploring decision architectures under uncertainty, Table \ref{tab:comparison-perspectives}.

All three cases demonstrated that the decision space can be substantially richer than the predefined strategies that serve as inputs to uncertainty-only analysis. The manufacturing case provides the most striking example: the existing valuation model was unable to support the decision because the company’s actual decision levers were not represented in the field-standard model.

Interesting insights emerged for the rugged landscape concept. Combining a structured decision space with external uncertainty can make the performance landscape even more rugged, as illustrated by the mortgage and M\&A cases. Moreover, exploration, rather than optimization, reframes the task from 'searching for the best strategy' to 'getting the most important choices right', as was specifically requested by the company in the manufacturing case.

The cases show that global sensitivity indices alone cannot provide the full picture. A choice that appears unimportant globally may matter within specific choice configurations, an observation made in every case. Regional sensitivity analysis can partly address this problem by quantifying variable importance within regions of the output. However, it can still disregard important relationships when the configuration in which a choice matters overlaps in performance with other configurations in which it does not. More generally, sensitivity indices do not reveal the direction or shape of input-output relationships. For this reason, complementing global sensitivity analysis with exploratory visualization of conditional distributions becomes critical for strategy design.

All three cases demonstrate that the notion of the best strategy is conditional on: uncertainty, other choices, and the desired performance characteristics that can emerge during the analysis. In some settings, low variance may be more attractive than a higher mean; in others, robustness across uncertainty scenarios may dominate expected performance in one scenario. The M\&A case adds a further insight: some choices are effectively real options, but optimizing these options in isolation would hide their nonlinear performance across choice configurations and uncertainty states.    

\begin{table}[htbp]
\centering
\caption{How the cases extend established decision-support perspectives}
\label{tab:comparison-perspectives}

\footnotesize
\setlength{\tabcolsep}{4pt}
\renewcommand{\arraystretch}{1.2}

\begin{tabularx}{\textwidth}{
p{1.9cm}
X
X
X
X}
\toprule
\textbf{Case} & 
\textbf{Uncertainty-only analysis} & 
\textbf{Rugged landscape} & 
\textbf{Global sensitivity analysis} & 
\textbf{Optimization} \\
\midrule

Mortgage 
& Limits attention to the default decision configuration 
& Even independent choices can change their relative performance under uncertainty 
& A choice with a near-zero sensitivity index may still matter if uncertainty explains most variance 
& The best-performing configuration is not necessarily the most robust across uncertainty \\

\addlinespace

Manufacturing upgrade 
& Cannot represent the decision space, as each uncertain model input is influenced by multiple alternative actions. 
& Focus on partial configurations prioritizes the most critical choices first 
& A choice with high sensitivity is not necessarily advisable if it has a negative effect or its positive effect overlaps with another influential choice; a low-sensitivity choice can still be critical locally (within an appealing partial configuration) 
& The best-performing configuration is implementing all upgrades, while the practical decision problem is prioritizing a smaller subset that creates most of the improvement \\

\addlinespace

M\&A 
& Limits attention to a single strategy
& Same choice configurations can perform differently in different regions of uncertainty
& Choices may earn low sensitivity indices if they only make a difference in a certain configuration of other choices
& Optimality of a learning threshold may depend on uncertainty and choice configuration \\

\bottomrule
\end{tabularx}
\end{table}


Conventional uncertainty-only analysis is often met with skepticism by practitioners, a pattern we observed during the development of both the manufacturing and M\&A cases. The common rationale is the garbage-in-garbage-out principle: if the numerical assumptions are questionable, the resulting decision recommendation is equally questionable. Indeed, the outcome of the uncertainty-only analysis is often a crisp decision threshold, such as a maximum acceptable probability of losses. Such a number can be highly sensitive to the underlying assumptions. 

Multichoice modeling shifts the focus from evaluating a single decision under uncertainty against a numerical threshold to exploring patterns of system behavior. As a result, the analysis becomes less dependent on any particular numerical outcome and more concerned with structural relationships, such as which choices matter, how they jointly shape the performance, and how they interact with uncertainty. These patterns are often driven by the structure and mechanics of the model rather than by the precise parameter values alone. Moreover, the identified inflection points can be further examined, challenged, or refined through targeted information gathering and expert elicitation.

In each case, the central question shifted from “How profitable is this strategy under uncertainty?” to “Which choices matter under which conditions, and which configurations of these choices form attractive strategies?” Accordingly, the purpose of the analysis shifted from evaluating a predefined strategy to exploring the space of available choices and designing a strategy from them. This observation highlights the importance of how decision situations are framed and the role analytical tools play in shaping that framing \citep{quade1980pitfalls}. It echoes the quotation that opened this paper:

\begin{quote}
\textit{“We shape our tools and thereafter our tools shape us.”}
\end{quote}

Various authors, commonly attributed to Marshall McLuhan \citep{Alex_Kusky}.

The three cases suggest that many established decision-support approaches implicitly constrain attention to a predefined set of alternatives, variables, or optimal solutions. By making the decision architecture explicit and computationally explorable, multichoice modeling shifts the focus toward discovering, comparing, and designing alternative configurations. In this sense, it expands not only the analytical space available to the decision maker, but also the range of strategies that can be meaningfully considered.

\section{Conclusion}

Decision-support approaches have become increasingly sophisticated in representing uncertainty, yet they often operate with a restricted representation of the decision space, typically evaluating one or a small number of predefined strategies. Strategy research, by contrast, has long recognized that strategic performance emerges from rugged landscapes of multiple interdependent choices. This paper connects these perspectives by conceptualizing decision support as the exploration of an uncertainty-contingent rugged decision landscape and offering a practical workflow to investigate joint uncertainty-decision space.

The three cases demonstrate how the focus of analysis shifts from the conventional valuation of a predefined strategy to the design and discovery of strategies from a richer decision space. Forcing the explicit representation of choices available to the decision-maker enlarges the set of strategies considered and can lead to configurations that differ materially from the predefined default. 

Future research could develop more systematic procedures for eliciting choice architectures, adaptive sampling designs for large configuration spaces, and applications in settings where strategic choices and uncertainty are deeply intertwined.


\bibliographystyle{informs2014} 
\bibliography{references} 





\end{document}